\documentclass[conference]{IEEEtran}
\IEEEoverridecommandlockouts
\usepackage{cite}
\usepackage{amsmath,amssymb,amsfonts}
\usepackage{algorithmic}
\usepackage{graphicx}
\usepackage{textcomp}
\usepackage{xcolor}

\usepackage{algorithm}
\usepackage{algorithmic}

\usepackage{multirow}
\usepackage{multicol}

\def\BibTeX{{\rm B\kern-.05em{\sc i\kern-.025em b}\kern-.08em
    T\kern-.1667em\lower.7ex\hbox{E}\kern-.125emX}}
\begin{document}

\title{
SRNeRV: A Scale-wise Recursive Framework for Neural Video Representation
}

\author{
}

\author{Jia Wang, Jun Zhu and Xinfeng Zhang~\IEEEmembership{Senior Member,~IEEE}
\thanks{

\textit{Corresponding author: Xinfeng Zhang.}

This work was supported by the National Natural Science Foundation of China (Grant No. 62461160310, 62521007, 62431011), the Fundamental Research Funds for the Central Universities (E2ET1104).

The first two authors contributed equally to this paper.
Jia Wang, Jun Zhu and Xinfeng Zhang are with School of Computer Science and Technology, University of Chinese Academy of Sciences, Beijing, China (E-mail: \{wangjia242, zhujun23\}@mails.ucas.ac.cn; xfzhang@ucas.ac.cn). 
}}

\maketitle

\begin{abstract}
Implicit Neural Representations (INRs) have emerged as a promising paradigm for video representation and compression. However, existing multi-scale INR generators often suffer from significant parameter redundancy by stacking independent processing blocks for each scale. Inspired by the principle of scale self-similarity in the generation process, we propose SRNeRV, a novel scale-wise recursive framework that replaces this stacked design with a parameter-efficient shared architecture. The core of our approach is a hybrid sharing scheme derived from decoupling the processing block into a scale-specific spatial mixing module and a scale-invariant channel mixing module. We recursively apply the same shared channel mixing module, which contains the majority of the parameters, across all scales, significantly reducing the model size while preserving the crucial capacity to learn scale-specific spatial patterns. Extensive experiments demonstrate that SRNeRV achieves a significant rate-distortion performance boost, especially in INR-friendly scenarios, validating that our sharing scheme successfully amplifies the core strengths of the INR paradigm.
\end{abstract}


\section{Introduction}

Video forms the largest source of internet traffic, making efficient compression essential for modern digital communication. Traditionally, this has been met by internationally standardized codecs~\cite{H.264, H.265, H.266} like H.265/HEVC~\cite{H.265}, which rely on a handcrafted pipeline of block transforms and motion compensation. The recent rise of deep learning, however, has introduced a powerful, data-driven alternative. This has driven an evolution in learned compression, moving from hybrid models that enhance parts of this traditional pipeline~\cite{li2018fully, pfaff2018neural, yan2017convolutional, zhang2017effective} to fully end-to-end frameworks like DCVC that learn the entire process from data~\cite{lu2019dvc, hu2021fvc, li2021dcvc}.

Prevailing end-to-end frameworks learn an amortized autoencoder on a large dataset, compressing any visual signal by mapping it to a compact latent space~\cite{kingma2013auto, kim2018semi}. In a conceptual shift, Implicit Neural Representations (INRs) create a functional representation unique to a single data instance, where the signal is embodied by the network parameters themselves~\cite{mildenhall2021nerf, sitzmann2020implicit, dupont2021coin, chen2021nerv}. This approach leverages the inherent smoothness between pixels or frames to model the signal as a continuous function, while simultaneously exploiting the repetitive computational logic found in the generative mapping from coordinate to value. A single, unified network serves as this function, encapsulating the shared logic while modulating its output based on the specific input coordinate. This paradigm of representing a signal as a learned, continuous function offers a new path towards highly compact representations.

\begin{figure}[t]
\centering
\includegraphics[width=0.5\textwidth, trim=0cm 13.5cm 24cm 0cm, clip]{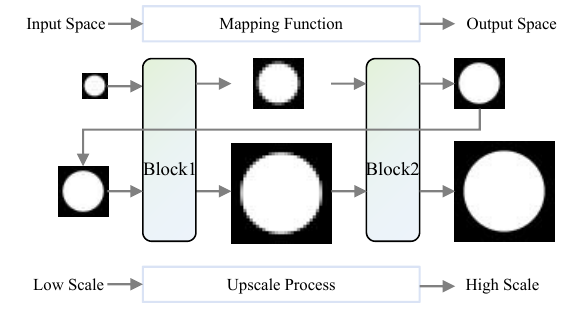} 
\caption{Self-similarity in the multi-scale feature and corresponding generation process.}
\label{fig:fig1}
\vspace{-10pt}
\end{figure}

\begin{figure*}[t]
\centering
\includegraphics[width=1.0\textwidth, trim=0cm 8cm 6cm 0cm, clip]{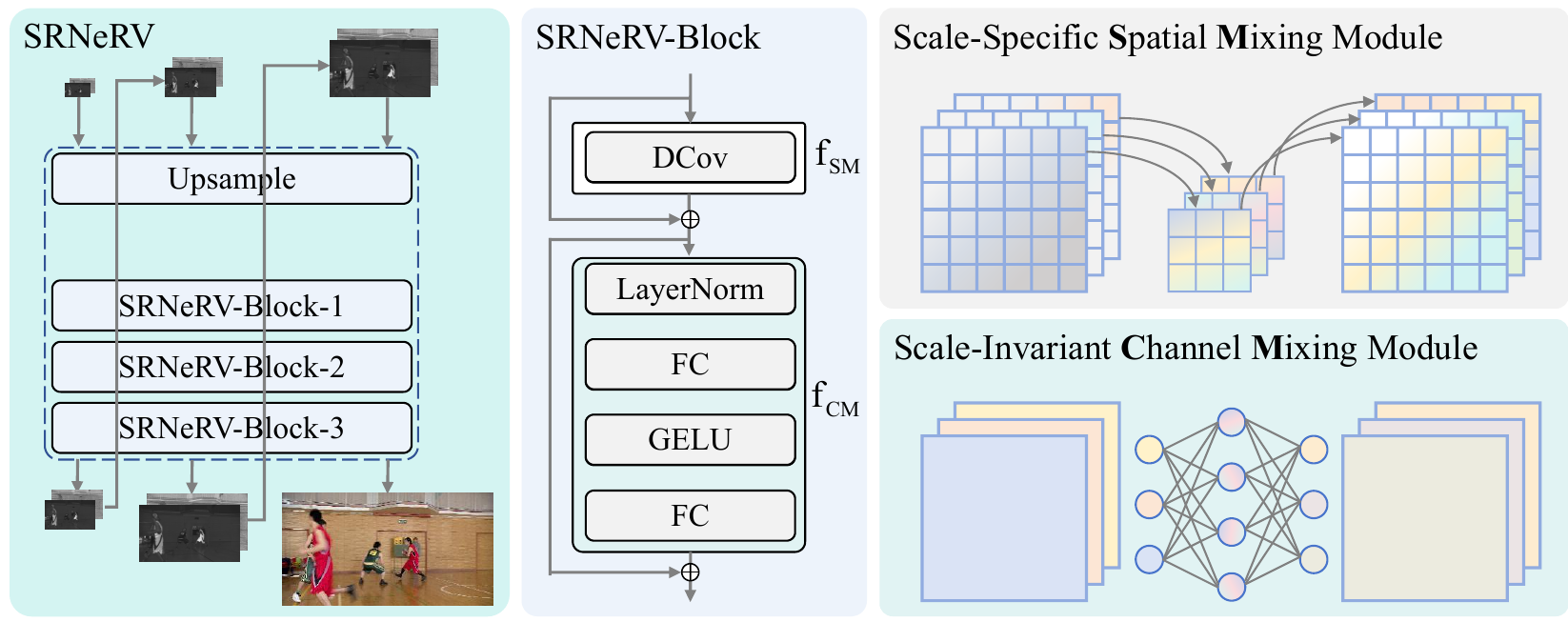} 
\caption{Overview of the SRNeRV architecture. (Left) The macro-architecture progressively generates high-scale frames by recursively applying shared SRNeRV-Blocks. (Center) The block's micro-architecture. (Right) Our hybrid sharing design is motivated by the distinct functions and parameter distribution of its components: the channel mixing Module contains most of parameters and performs a scale-invariant task, thus is the part shared across all scales.}
\label{fig:fig2}
\vspace{-5pt}
\end{figure*}

Just as INR is efficient due to shared logic across spatial coordinates, our key insight is that, across different scales, the generative mapping from lower to higher-resolution features is also a conceptually repetitive process. This self-similarity is illustrated in Fig.~\ref{fig:fig1}. The figure shows a process where the complete upsampling step, which transforms a low-scale input into a refined high-scale output (realized by the combined function of Block1 and Block2), is fundamentally alike at each stage. Such repetition suggests that a single, unified upsampling block could be applied recursively to perform this task. Indeed, this principle of leveraging scale self-similarity with a shared functional block is a cornerstone of classical computer vision, most famously demonstrated by the Laplacian pyramid~\cite{burt1987laplacian}.

Building on this principle of scale self-similarity, we propose a novel scale-wise recursive framework for INR-based video representation, termed SRNeRV. The core design of SRNeRV decouples the processing block into two functional components: a scale-specific spatial mixing module and a scale-invariant channel mixing module. We then introduce a hybrid sharing scheme where only the channel mixing module is shared across all scales. This shared module is applied recursively: the output features from one scale become the input for the module at the next scale, enabling the progressive generation of high-scale, refined features. This hybrid approach allows the model to learn unique spatial filters for each resolution level, thereby maintaining the capacity to capture scale-specific patterns, while drastically reducing the total parameter count by sharing the much larger channel mixing Feedforward Network (FFN).

The main contributions of our work are summarized as follows:
\begin{enumerate}
\item To the best of our knowledge, we are the first to systematically analyze and exploit the scale self-similarity inherent in the INR generation process, extending the core principle of INR from coordinate-wise logic to multi-scale generative logic.
\item We propose SRNeRV, a novel and highly compact scale-wise recursive framework for video representation based on a hybrid parameter sharing scheme that decouples spatial and channel mixing.
\item We conduct extensive experiments on standard video sequences, demonstrating the better rate-distortion performance of our model and validating the effectiveness of our proposed hybrid recursive design.
\end{enumerate}

\section{Related Works}
\textbf{Implicit Neural Representations.}
INRs are a powerful paradigm that represents a signal as a continuous function, $f(coord) \to value$, parameterized by a neural network. This functional representation paradigm has proven to be highly flexible. It has been successfully applied to represent 2D images by mapping spatial coordinates to pixel values~\cite{sitzmann2020implicit, dupont2021coin}, 3D scenes by mapping viewing directions to color and density~\cite{mildenhall2021nerf}, and entire videos by mapping time indices to frames~\cite{chen2021nerv, kwan2024hinerv, wang2025uar, zhang2024vqnerv, tang2025canerv}. Our work extends this principle by applying it internally within the synthesis network: we learn a function that maps low-scale features to high-scale features.

\textbf{Iterative Refinement via Shared Functions.}
The principle of iterative refinement via a shared function has a rich history, from the classic Laplacian Pyramid~\cite{burt1987laplacian} that recursively applies a fixed kernel, to early deep recursive networks like DRCN~\cite{kim2016deeply} that repeatedly apply a learned block for image restoration. 
This concept is central to modern generative modeling. A prime example is the Denoising Diffusion Probabilistic Model (DDPM)~\cite{ho2020denoising}, which transforms noise into an image by iteratively applying a single, shared network across hundreds of steps.
Our work applies this powerful principle to a new domain: the spatial scale axis of an INR's synthesis network, using a shared block to recursively generate higher-scale features.

\section{Methods}

\subsection{Preliminaries}
Modern high-performance INRs for video typically employ a multi-scale synthesis network. This network architecture begins with a low-dimensional input, such as an encoded time index $t$ or a learnable grid embedding, and progressively generates the final high-resolution frame through a series of $M$ upsampling stages.
At each stage $i$, the input feature map from the previous stage, $x_{i-1}$, is first spatially enlarged by an upsampling layer. This is followed by a stack of $L$ feature refinement blocks to process and enhance the features at the new scale:
\begin{equation}
  h_i = \text{Upsample}(x_{i-1}),
\end{equation}
\begin{equation}
  x_i = \text{RefinementBlocks}_i(h_i).
\end{equation}

A modern refinement block, such as the ConvNeXt block used in state-of-the-art models~\cite{kwan2024hinerv}, typically decouples its operation into two functional components: a spatial mixing module (e.g., a depthwise convolution) to aggregate local information, and a channel mixing module (e.g., an FFN) to perform feature transformation.

In this conventional stacked design, each refinement block at each stage is instantiated with a unique set of parameters. This design philosophy overlooks the inherent self-similarity of the refinement task across different scales, leading to significant parameter redundancy and motivating our recursive approach.

\subsection{Our Framework}

Building on the principle of scale self-similarity, we introduce SRNeRV, a recursive framework designed for parameter efficiency as shown in Fig.~\ref{fig:fig2}. The core of our framework is a novel hybrid parameter sharing scheme applied to the refinement blocks.

\textbf{Overall architecture.}
The overall architecture of SRNeRV follows a recursive, coarse-to-fine generation process. Starting from an initial feature grid $x_0$, the network progressively generates features at higher scales. The entire process for M upsampling stages is outlined in Algorithm~\ref{alg:rnerv}.
At each upsampling stage $i$, a sequence of $L$ SRNeRV-Blocks is applied. The key innovation lies in how the parameters for these blocks are structured and shared.

\begin{algorithm}[t]
\caption{SRNeRV Generation Process}
\label{alg:rnerv}
\begin{algorithmic}[1] 
\STATE \textbf{Input:} Initial feature grid $\mathbf{x}_0$
\STATE \textbf{Parameters:}
\STATE \quad Spatial Mixing weights $\{ \theta_{\text{SM}_{i,j}} \}$ for $i \in \{1, \dots, M\}, j \in \{1, \dots, L\}$
\STATE \quad Channel Mixing weights $\{ \theta_{\text{CM}_j} \}$ for $j \in \{1, \dots, L\}$
\STATE
\STATE $\mathbf{x}_{\text{in}} \gets \mathbf{x}_0$
\FOR{$i = 1$ to $M$}
    \STATE $\mathbf{h} \gets \text{Upsample}(\mathbf{x}_{\text{in}})$
    \FOR{$j = 1$ to $L$}
        \STATE $\mathbf{y} \gets f_{\text{SM}}(\mathbf{h}; \theta_{\text{SM}_{i,j}}) + \mathbf{h}$
        \STATE $\mathbf{h} \gets f_{\text{CM}}(\mathbf{y}; \theta_{\text{CM}_j}) + \mathbf{y}$
    \ENDFOR
    \STATE $\mathbf{x}_{\text{in}} \gets \mathbf{h}$
\ENDFOR
\STATE \textbf{return} $\mathbf{x}_{\text{in}}$
\end{algorithmic}
\end{algorithm}

\textbf{Hybrid sharing in SRNeRV-Block.}
The SRNeRV-Block decouples the refinement operation into two distinct functional modules: a Scale-Specific Spatial Mixing module $f_{SM}$ and a Scale-Invariant Channel Mixing module $f_{CM}$.
Given an input tensor $\mathbf{h}$, the forward pass of a SRNeRV-Block is defined as:
\begin{equation}
\mathbf{y} = f_{SM}(\mathbf{h}; \theta_{SM_{i,j}}) + \mathbf{h},
\end{equation}
\begin{equation}
\mathbf{z} = f_{CM}(\mathbf{y}; \theta_{CM_j}) + \mathbf{y}.
\end{equation}

The Scale-Specific Spatial Mixing module, $f_{SM}$, is realized by a depthwise convolution. Its parameters, $\theta_{SM_{i,j}}$, are unique for each block, indexed by both the stage $i$ and the intra-stage position $j$. This allows the network to learn spatial patterns that are specific to each scale and processing depth.

In contrast, the Scale-Invariant Channel Mixing module, $f_{CM}$, is realized by an FFN. Its parameters, $\theta_{CM_j}$, are shared across all upsampling stages $i$ and only depend on the intra-stage position $j$. This design is based on the insight that the abstract logic of channel-wise feature transformation is reusable across different scales.
This hybrid sharing scheme is highly parameter-efficient, as the $f_{CM}$ modules, which are shared, constitute the vast majority of the parameters.

\subsection{Compression Pipeline}
We implement our SRNeRV framework by adapting the HiNeRV~\cite{kwan2024hinerv} architecture. Specifically, we replace the original independent refinement blocks, HiNeRV blocks, with our proposed SRNeRV-Blocks, which operate under the hybrid sharing scheme shown in Fig.~\ref{fig:fig2}. The compression pipeline follows the established per-instance fitting paradigm. 

The process begins by training SRNeRV to fit the video sequence. This is followed by a short phase of Quantization-Aware Training (QAT) to adapt the model to a lower bit precision. Finally, the converged quantized weights are serialized and losslessly compressed using an arithmetic coder to generate the final bitstream. The total bitrate of network $R$ is the sum of the codelengths of the scale-specific spatial mixing parameters $\theta_{\text{SM}}$, and the shared channel mixing parameters $\theta_{\text{CM}}$:
\begin{equation}
  R =  \sum_{i=1}^{M}\sum_{j=1}^{L}  - \log_2(p(\theta_{\text{SM}_{i,j}}))
       + \sum_{j=1}^{L}  - \log_2(p( \theta_{\text{CM}_j})), 
\end{equation}
where $p(\theta)$ is the estimated probability of each quantized weight value. Our primary contribution is the parameter-efficient architecture that minimizes the second term through sharing, while this compression backend remains identical to our baseline.

\section{Experiments}

\begin{table*}[t]
\centering
\begin{tabular}{c|cc|cc|ccccc|ccc}
\hline
\multirow{2}{*}{Methods} & \multicolumn{2}{c|}{SCC} & \multicolumn{2}{c|}{HEVC ClassE} & \multicolumn{5}{c|}{HEVC ClassB} & \multicolumn{3}{c}{UVG} \\
 & SCC-1 & SCC-2 & Four. & Kristen. & BQTerrace & Basket. & Cactus & Kimono & Park. & Bospho. & Jockey & Ready. \\ \hline
HNeRV (CVPR23) & -29.1 & -50.1 & 66.8 & 50.4 & 76.7 & 61.0 & 44.6 & 89.8 & 68.4 & 30.1 & 82.8 & 91.9 \\
Boost-NeRV (CVPR24) & -52.7 & -66.1 & -24.7 & -49.9 & 30.1 & 13.9 & -30.2 & 56.26 & 30.34 & 15.6 & 60.7 & 72.2 \\
HiNeRV (NIPS23) & -64.1 & -73.8 & -46.9 & -67.8 & -39.8 & -8.1 & -47.7 & 40.6 & -22.4 & -69.1 & 10.1 & 9.6 \\
SRNeRV-FullShare & -77.2 & -82.9 & -58.0 & -69.3 & -62.3 & -6.1 & -49.6 & 42.2 & -26.8 & -66.4 & 5.8 & 4.9 \\
SRNeRV & -84.0 & -87.7 & -62.5 & -77.8 & -67.4 & -14.3 & -52.8 & 38.9 & -29.6 & -72.3 & -1.7 & -6.1 \\ \hline
\end{tabular}
\vspace{3pt}
\caption{The BDBR(\%) performances of different settings when compared with H.266 on 12 videos.
}
\vspace{-10pt}
\label{table:BDBR}
\end{table*}

\begin{figure*}[t]
\centering
\includegraphics[width=\linewidth]{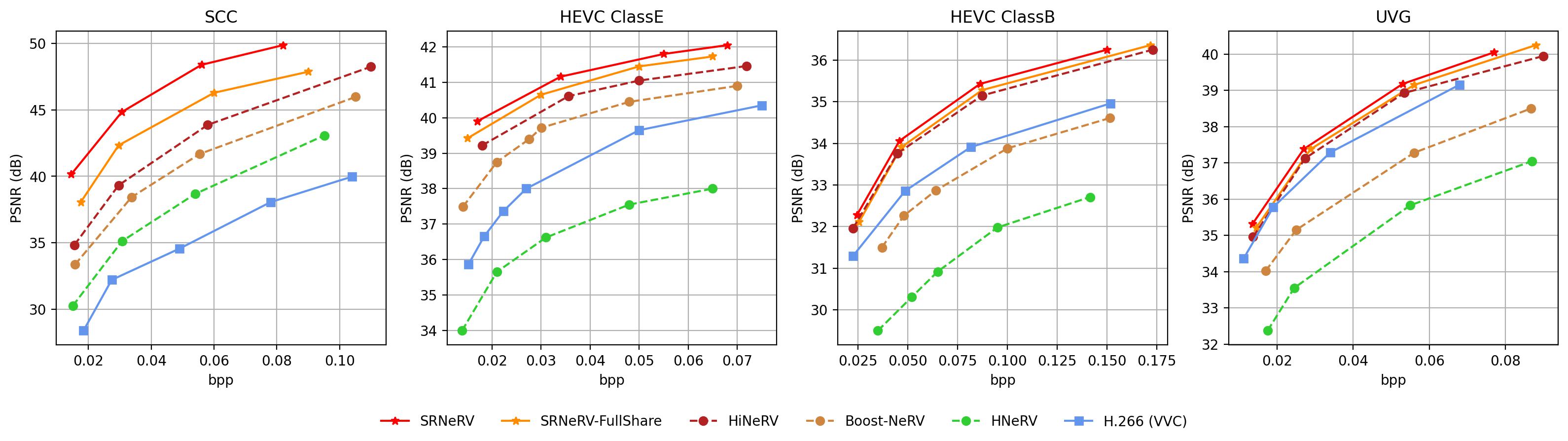}
\caption{Rate Distortion (RD) performance on different datasets.}
\label{fig:rd}
\end{figure*}

\subsection{Implementation}
\textit{1) Datasets}: To rigorously validate our framework, we conduct experiments across a broad and challenging range of video content. Our evaluation suite includes:
The \textbf{UVG} dataset~\cite{mercat2020uvg}, which is a standard benchmark for previous video compression methods~\cite{chen2021nerv, chen2023hnerv, kwan2024hinerv, zhang2024boosting} due to its high-resolution natural scenes.
The complete \textbf{HEVC Class B} collection, serving as a standard for evaluating performance on challenging, high-definition video.
The \textbf{HEVC Class E} set, which is characterized by complex foreground motion against a relatively static or simple background.
Two representative \textbf{Screen Content Coding (SCC)} sequences, which serve as a stress test for the model's ability to handle high-frequency details like text and graphics.
This diverse selection ensures that our performance claims are well-supported across various content types, from common INR benchmarks to specialized and challenging video categories.

\textit{2) Comparison Methods}: We benchmark SRNeRV against a comprehensive suite of baselines. Traditional Codecs: We include the H.266/VVC. INR-based Codecs: Our main baselines are HNeRV\cite{chen2023hnerv}, Boost-NeRV~\cite{zhang2024boosting} (using its best performance version HNeRV-Boost), and HiNeRV~\cite{kwan2024hinerv}. 
Ablation Variant: We include SRNeRV-FullShare which shares the whole SRNeRV-Block across scale.

\textit{3) Implementation Details}: For a fair comparison, SRNeRV is implemented within the official HiNeRV codebase and trained from scratch for each sequence. Unless specified otherwise, models are trained for 300 epochs using the Adam optimizer with a 2e-3 learning rate and a cosine annealing schedule. The loss is a sum of L1 and MS-SSIM loss following previous works~\cite{ kwan2024hinerv}. We follow the established compression pipeline, including a 15\% weight pruning step and an 6-bit QAT stage. To generate Rate-Distortion (RD) curves, model scales for all INR methods are configured to ensure comparable parameter counts.

\textit{4) Evaluation Metrics}: We use PSNR to measure reconstruction quality and bits per pixel (bpp) for bitrate. Overall RD performance is compared using the Bjontegaard Delta Bit-Rate (BDBR)~\cite{bjontegaard2001BDBR}.

\subsection{Comparison Results}
\textbf{Overall performance.} We present the main quantitative results in Table~\ref{table:BDBR} and Fig.~\ref{fig:rd}. Across all tested datasets, from UVG to challenging HEVC Class B/E and SCC sequences, our proposed SRNeRV consistently outperforms its direct predecessor, HiNeRV, as well as other strong INR baselines. This demonstrates the broad applicability and superior parameter efficiency of our hybrid recursive design.

\textbf{Analysis of performance gains.} While SRNeRV demonstrates broad performance gains, the improvements are most significant on sequences where INRs traditionally excel: those with temporally redundant or simple backgrounds, such as the HEVC Class E and SCC sets. This pronounced advantage in INR-friendly scenarios validates that \textbf{our hybrid sharing scheme successfully amplifies the core strengths of the INR paradigm.} By sharing the large FFN to compactly model the static background, more of the parameter budget is freed for the scale-specific spatial modules to capture complex foreground details, from the sharp graphics in SCC to the intricate motion in Class E.

\textbf{Ablation on sharing strategy.} The performance of SRNeRV and SRNeRV-FullShare confirms our central hypothesis. Even a naive full-sharing approach improves upon the non-sharing baseline method HiNeRV, validating the general benefit of exploiting scale self-similarity. Crucially, our proposed hybrid-sharing SRNeRV achieves a significantly larger gain, demonstrating that retaining scale-specific spatial modules is key to balancing parameter compactness with high-fidelity reconstruction.

\section{Conclusion}
In this paper, we introduced SRNeRV, a scale-wise recursive framework that addresses parameter redundancy in multi-scale INR generators via a novel hybrid parameter sharing scheme. By decoupling and selectively sharing the scale-invariant channel mixing module, SRNeRV achieves a highly compact representation while retaining crucial scale-specific processing capabilities. Our experimental results validate this design, demonstrating not only a better overall rate-distortion performance but also a pronounced advantage in INR-friendly scenarios like screen content. This confirms that our approach successfully amplifies the core representational strengths of the INR paradigm. We believe this principle of targeted recursive sharing offers a promising direction for future efficient neural representation designs.

\bibliographystyle{IEEEtran}   
\bibliography{refs}            


\end{document}